\documentclass[letterpaper, 10 pt, conference]{ieeeconf}  

\IEEEoverridecommandlockouts                              

\usepackage{graphics} 
\usepackage{epsfig} 
\usepackage{mathptmx} 
\usepackage{times} 
\usepackage{amsmath} 
\usepackage{amssymb}  
\usepackage{cite}
\usepackage{etoolbox}
\makeatletter
\patchcmd{\@makecaption}
  {\scshape}
  {}
  {}
  {}
\makeatletter
\patchcmd{\@makecaption}
  {\\}
  {.\ }
  {}
  {}
\makeatother

\title{\LARGE \bf
RASNet: Segmentation for Tracking Surgical Instruments in Surgical Videos Using Refined Attention Segmentation Network
}

\author{Zhen-Liang Ni$^{1,3}$, Gui-Bin Bian$^{1,3}$, Xiao-Liang Xie$^{1}$, Zeng-Guang Hou$^{1,2,3}$
\\Xiao-Hu Zhou$^{1,3}$  and Yan-Jie Zhou$^{1,3}$  
\thanks{$^{1}$State Key Laboratory of Management and Control for Complex Systems,
Institute of Automation, Chinese Academy of Sciences, Beijing 100190,
China.}%
\thanks{$^{2}$CAS Center for Excellence in Brain Science and Intelligence Technology, Beijing 100190, China}
\thanks{$^{3}$University of Chinese Academy of Sciences, Beijing 100049, China.}
\thanks{Email: $\{$nizhenliang2017, guibin.bian, xiaoliang.xie, zengguang.hou, zhouxiaohu2014, zhouyanjie2017$\}$ @ia.ac.cn}
}

\begin{document}

\maketitle
\thispagestyle{empty}
\pagestyle{empty}

\begin{abstract}
Segmentation for tracking surgical instruments plays an important role in robot-assisted surgery. Segmentation of surgical instruments contributes to capturing accurate spatial information for tracking. In this paper, a novel network, Refined Attention Segmentation Network, is proposed to simultaneously segment surgical instruments and identify their categories. The U-shape network which is popular in segmentation is used. Different from previous work, an attention module is adopted to help the network focus on key regions, which can improve segmentation accuracy. To solve the class imbalance problem, the weighted sum of the cross entropy loss and the logarithm of the Jaccard index is used as the loss function. Furthermore, transfer learning is adopted in our network. The encoder is pre-trained on ImageNet. The dataset from the MICCAI EndoVis Challenge 2017 is used to evaluate our network. Based on this dataset, our network achieves state-of-the-art performance 94.65$\%$ mean Dice and 90.33$\%$ mean IOU.

\end{abstract}

\section{Introduction}
With the increasing popularity of robot-assisted surgery, tracking and segmentation of surgical instruments have lately received great attention because of their promising applications in robot-assisted surgery. Its potential applications in surgery include precise localization and pose estimation of surgical instruments, real-time reminders of surgery, surgical workflow optimation, objective skill assessment, etc~\cite{a1,a2}. MICCAI held Robotic Instrument Segmentation sub-challenge in the MICCAI EndoVis Challenge 2017~\cite{a6}. They also held Robotic Scene Segmentation sub-challenge in the MICCAI EndoVis Challenge 2018. These challenges have attracted much attention to the research of surgical instrument segmentation.

A great deal of research has been done on tracking surgical instruments. Bareum et al~\cite{a3} modified YOLO for real-time tracking of surgical instruments. Amy et al~\cite{a4} applied Faster R-CNN to the classification of surgical instruments and bounding box regression. Duygu et al~\cite{a1} merged region proposal network and a two-stream convolutional network to detect surgical instruments. However, most of these work can only obtain the bounding box but not the precise boundaries of surgical instruments. Precise boundary of surgical instruments is essential to robot-assisted surgery. Iro et al~\cite{a5} proposed a new convolutional neural network called CSL model to simultaneously perform segmentation and pose estimation of surgical instruments. This work provides a new direction for tracking surgical instruments. Inspired by Iro~\cite{a5}, we propose a new network to segment and classify surgical instruments for tracking.

Segmentation of surgical instruments faces many challenges. Due to the narrow field of view, sometimes only a part of the surgical instrument appears in the field of view. Besides, when the pose of the surgical instrument changes, the geometry appearing in the field of view is different. To overcome these issues, the network must be able to capture deep semantic features. Another important issue is the class imbalance issue. Surgical instruments usually only occupy a small region of the entire field of view. The number of foreground pixels is much smaller than that of background pixels, which leads to serious class imbalance.

To overcome the issues mentioned above, a novel network, Refined Attention Segmentation Network(RASNet), is proposed which utilizes attention mechanism for semantic segmentation of surgical instruments. Attention mechanism can help the network capture deeper semantic features by utilizing the global context of high-level features~\cite{a7,a8,a9}. The U-shape network is outstanding in making full use of the detail information in the low-level features and the semantic information in the high-level features~\cite{a11}. The combination of the cross entropy loss and the logarithm of the Jaccard index is used as the loss function to solve the class imbalance problem and improve segmentation performance for small objects~\cite{a13}. Furthermore, transfer learning is adopted in our network to improve segmentation accuracy.
\section{Methods}
\begin{figure*}[thpb]
  \centering
  \includegraphics[width=0.96\textwidth]{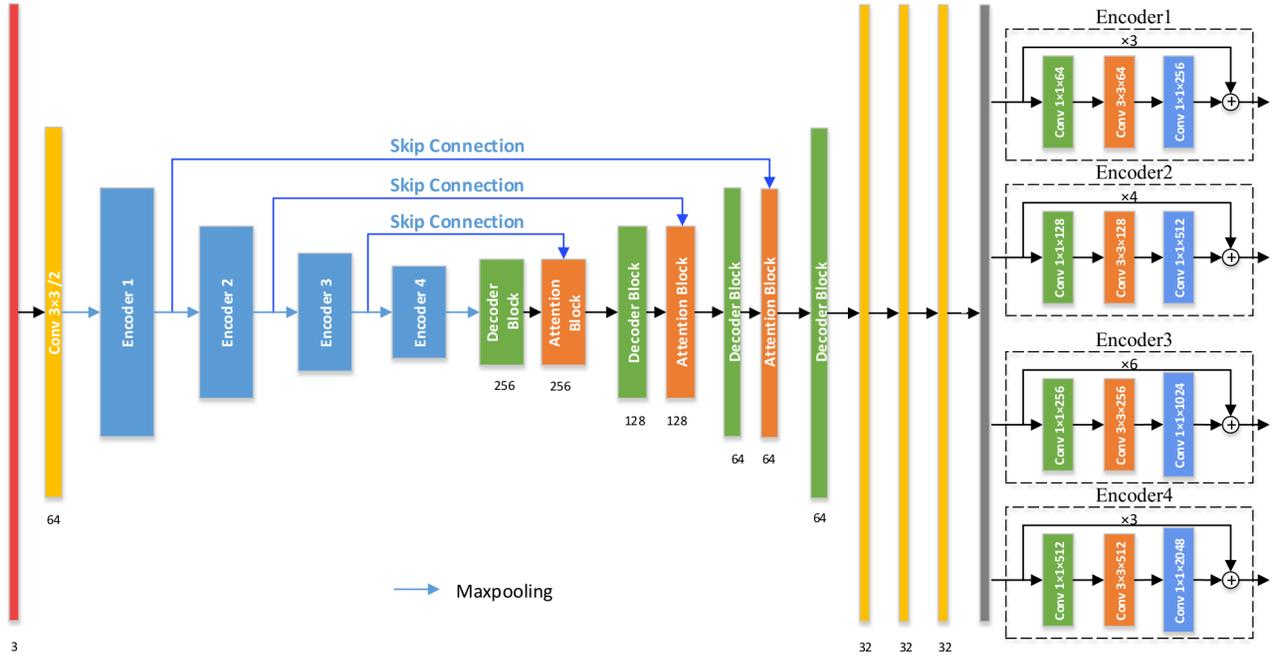}
  \caption{The Architecture of Refined Attention Segmentation Network. RASNet is a U-shape network which consists of a contracting path and an expanding path. ResNet-50 is used as the encoder. The decoder consists of Attention Fusion Module and Decoder Block.}
  \label{fig1}
\end{figure*}

\subsection{Network Architecture}
The architecture of RASNet is presented in Fig.~\ref{fig1}. RASNet consists of a contracting path and an expanding path. The contracting path is used to capture deep semantic features, and the expanding path is used for precise localization~\cite{a11}. RASNet uses ResNet-50~\cite{a12} pre-trained on ImageNet as its encoder. The encoder starts with a convolution layer with a kernel of size 7$\times$7 and a stride of 2. Then there is a max-pooling layer with a kernel of size 3$\times$3 and a stride of 2. The following portion of encoder includes 4 encoder blocks. Each encoder block consists of several residual blocks. The structure of four encoder blocks is presented in Fig.~\ref{fig1}. To reduce the loss of information, RASNet uses deconvolution for upsampling. RASNet merges the high-level feature maps with low-level feature maps via skip connections. Different from previous work, an attention module is designed to fuse high-level features with low-level features efficiently. The output of the network is a mask of the same size as the original image.
\subsection{Attention Fusion Module}
\begin{figure}[tbhp]
  \centering
  \includegraphics[width=0.46\textwidth]{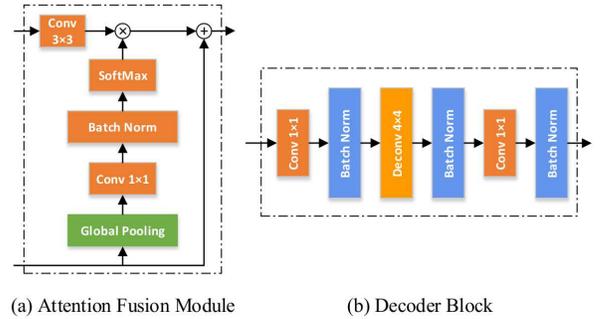}
  \caption{Components of Attention Fusion Module and Decoder Block.}
  \label{fig2}
\end{figure}

U-Net directly concatenates low-level features with high-level features. But that is a naive method. High-level features contain abundant semantic information which can help low-level features select precise location information. Therefore, we utilize an improved Global Attention Upsample module~\cite{a8} which can be called Attention Fusion Module$($AFM$)$. In AFM, the global average pooling is performed to extract the global context of high-level features which is represented as a vector. A 1$\times$1 convolution with batch normalization is used to normalize weights. Then a softmax function is used as a nonlinear activation function to make the sum of the weights equal to 1, which is different from the Global Attention Upsample module. The low-level features are multiplied by the weight vector. Then the weighted lower-level features are added to the higher-level features. This module fuse multi-level features effectively by utilizing the global context of high-level features as guidance information. The architecture of AFM is shown in Fig.~\ref{fig2}.

\subsection{Decoder Block}
The architecture of the decoder block is illustrated in Fig.~\ref{fig2}. It starts with a 1$\times$1 convolution with batch normalization to reduce the dimension of feature maps, which can cut down on computation complexity. Then a 4$\times$4 deconvolution is performed with a stride of 2 and batch normalization to upsample the feature maps. Finally, a 1$\times$1 convolution with batch normalization is performed to adjust the dimension of feature maps.
\subsection{Loss Function}
As loss function, we use the weighted sum of the cross entropy loss and the logarithm of the Jaccard index~\cite{a13}. The task of segmenting surgical instruments can be considered as a classification for pixels. So the cross entropy loss is chosen which is represented by $H$ in Equation 1.
$$
H =  - \frac{1}{{w \times h}}\sum\limits_{k = 1}^c {\sum\limits_{i = 1}^w {\sum\limits_{j = 1}^h {{y_{ijk}}} } \log (\frac{{{e^{{{\widehat y}_{ijk}}}}}}{{\sum\limits_{k = 1}^c {{e^{{{\widehat y}_{ijk}}}}} }})} \eqno{(1)}
$$
where $w$, $h$ represent the width and the height of the predictions. And c is the number of classes. $y_{ijk}$ is the ground truth of a pixel and ${\widehat y}_{ijk}$ is the prediction of a pixel.

In addition, small surgical instruments can cause class imbalance problem. The Jaccard index can reduce the impact of class imbalance on the network. The Jaccard index is represented by $J$ in Equation 2.
$$
J = \frac{{TP}}{{TP + FP + FN}} \eqno{(2)}
$$
where TP, FP and FN represent true positives subset of pixels, false positives subset of pixels and false negatives subset of pixels, respectively.

The loss function is shown in Equation 3.
$$
L = H - \alpha \log (J) \eqno{(3)}
$$
where $\alpha$ is a weight which can balance cross entropy loss and the logarithm of the Jaccard index.

\section{Experiments And Results}
\subsection{Dataset}
The dataset used in this paper is from the MICCAI EndoVis Challenge 2017~\cite{a6}. Since the labels of test set from the EndoVis Challenge 2017 are not obtained, we only use the training set from the EndoVis Challenge 2017 to evaluate our network. The training set from the EndoVis Challenge 2017 consists of eight sequences of high-resolution stereo camera images acquired from a da Vinci Xi surgical system during endoscopic surgery. It includes 1800 images with a resolution of 1920*1080. 400 frames are selected as the test set. In the selection process, we usually select consecutive sequences to avoid similar frames in the training set and test set. The remaining 1400 frames are used as the training set. Surgical instruments in the dataset can be divided into six categories, namely Bipolar Forceps, Prograsp Forceps, Needle Driver, Vessel Sealer, Grasping Retractor and Curved Scissors. A miscellaneous category is labeled for any other surgical instrument. The number of samples for each category is shown in Table~\ref{tab1}.
\begin{table}[htbp]
  \centering
  \caption{Details of Dataset}
    \begin{tabular}{cccc}
    \hline
    Instrument & Number of Instances & Training Set& Test Set\\
    \hline
    Bipolar Forceps & 657   & 445   & 212 \\
    Prograsp Forceps & 841   & 754   & 87 \\
    Needle Driver & 1197  & 1013  & 184 \\
    Vessel Sealer & 386   & 276   & 110 \\
    Grasping Retractor & 228   & 210   & 18 \\
     Curved Scissors & 351   & 302   & 49 \\
    Other & 449   & 422   & 27 \\
    Total & 4109  & 3422  & 687 \\
    Number of Frames & 1800  & 1400  & 400 \\
    \hline
    \end{tabular}%
  \label{tab1}%
\end{table}%

\subsection{Experimental Details}
ResNet-50 pre-trained on ImageNet is utilized to initialize the encoder. This initialization strategy helps improve network performance~\cite{a13}. The effectiveness of this initialization strategy is verified by comparing network performance with different initialization strategies. Then, the networks without AFM are tested and compared to RASNet to prove the effectiveness of AFM. U-Net and TernausNet are chosen to compare with RASNet~\cite{a11,a13}. Both of these networks have excellent performance in medical image segmentation. All networks are trained and tested based on the dataset illustrated in Table~\ref{tab1}. Intersection Over Union(IOU) and Dice ratio are selected as the evaluation metric, which all can serve as the similarity measurement between two sets.

Due to limited computing resources, all images used in the experiment are resized to 320$\times$256 pixels. The network is trained using Adam with batch size 8. The network is easy to overfit due to the pre-training of the encoder. To prevent network overfitting, a strategy to dynamically adjust the learning rate is adopted. The initial learning rate is $3 \times {10^{ - 4}}$. Learning rate multiplied by 0.8 every 30 iterations. After a series of experiments, we set the weight $\alpha$ in the loss function to 0.3.

\subsection{Results}
In the experiments, two initialization strategies with RASNet are tested. The performance of the two strategies is shown in Table~\ref{tab2}. The RASNet with random initialization achieves 78.11$\%$ mean Dice and 70.72$\%$ mean IOU. The network pre-trained gets 94.65$\%$ mean Dice and 90.33$\%$ mean IOU. These results prove that pre-training the encoder can greatly improve segmentation accuracy. This may be due to the fact that the ImageNet for pre-training contains rich visual features. The performance of the encoder to extract semantic features is improved by pre-training.
\begin{figure}[htpb]
  \centering
  \includegraphics[width=0.47\textwidth]{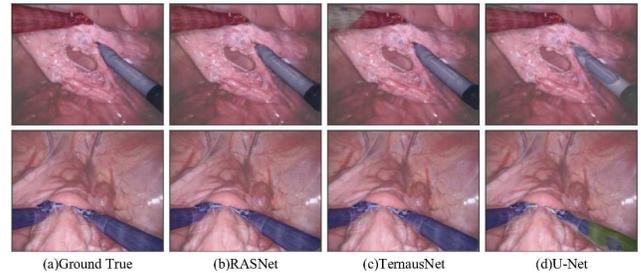}
  \caption{Example results of robotic instrument segmentation. Different types of instruments are distinguished by masks with different colors. Image (a) shows an original image with the ground true mask. Image (b) shows the prediction of RASNet which is proposed in this paper. Image (c) shows the prediction of TernausNet. And Image (d) shows the prediction of U-Net.}
  \label{fig4}
\end{figure}

\begin{table}
\caption{The performance comparison of RASNet with different initialization strategy}
\label{tab2}
\begin{center}
\begin{tabular}{c c c}
\hline
Method & Mean Dice($\%$) & Mean IOU($\%$)\\
\hline
Random initialization & 78.11 & 70.72\\
Pre-trained & 94.65 & 90.33\\
\hline
\end{tabular}
\end{center}
\end{table}

To evaluate the performance of AFM, two networks are tested. One with AFM and the other one without it. Table~\ref{tab3} shows the results of the experiments. As shown in Table~\ref{tab3}, the network with AFM achieves 94.65$\%$ mean Dice and 90.33$\%$ mean IOU. The network without it just gets 89.31$\%$ mean Dice and 82.75$\%$ mean IOU. Network performance is increased by 5.34$\%$ Dice and 7.58$\%$ IOU by using AFM.
\begin{table}[htbp]
\caption{Evaluation results of different networks.} 
\label{tab3}
\begin{center}
\begin{tabular}{c c c}
\hline
Method & Mean Dice($\%$) & Mean IOU($\%$)\\
\hline
RASNet without AFM & 89.31 & 82.75\\
RASNet & 94.65 & 90.33\\
\hline
U-Net & 70.04 & 56.76\\
TernausNet & 88.04 & 80.34\\
\hline
\end{tabular}
\end{center}
\end{table}

\begin{table}[htbp]
  \centering
  \caption{Mean Dice and mean IOU of each class}
    \begin{tabular}{ccc}
    \hline
    Instrument & Mean Dice(\%) & Mean IOU(\%) \\
    \hline
    Bipolar Forceps & 97.75 & 95.61 \\
    Prograsp Forceps & 84.59 & 73.31 \\
    Needle Driver & 98.31 & 96.68 \\
    Vessel Sealer & 98.01 & 96.08 \\
    Grasping Retractor & 88.05 & 78.65 \\
    Curved Scissors & 99.01 & 98.04 \\
    Other & 96.87 & 93.95 \\
    Mean  & 94.65 & 90.33 \\
    \hline
    \end{tabular}%
  \label{tab4}%
\end{table}%

To further demonstrate the superior performance of our model, U-Net and TernuasNet are evaluated using the dataset illustrated in Table~\ref{tab1} to compare with RASNet. The results are presented in Table~\ref{tab3}. U-Net gets 70.04$\%$ mean Dice and 56.76$\%$ mean IOU. TernausNet achieves 88.04$\%$ mean Dice and 80.34$\%$ mean IOU. Their performance is much poor than RASNet. Also, to give a more intuitive comparison, we visualize the predictions of different models. The comparison results are presented in Fig.~\ref{fig4}. As Fig.~\ref{fig4} shows, the prediction of RASNet is the same as the ground truth. There is an error in the prediction of TernausNet that misclassify surgical instruments as background. The prediction of U-Net not only misclassifies surgical instruments as background but also confuses one surgical instrument with other types. In summary, RASNet achieves state-of-the-art performance in the segmentation of surgical instruments.

\section{Discussion}
Experimental results mentioned above confirm that RASNet achieves state-of-the-art performance in the segmentation of surgical instruments. The good performance of the RASNet is mainly attributed to the following three points. Resnet-50 Pre-trained is used to initialize the encoder. AFM extracts the global context of the high-level features to guide the low-level features to select precise position information. The loss merges the cross entropy and the logarithmic of the Jaccard index is used to solve the class imbalance problem.

Table~\ref{tab4} shows the mean Dice and mean IOU of each class. According to the results in Table~\ref{tab4}, Curved Scissors is the highest performing instrument, which may be due to its usage pattern and shape. Its shape is easy to be distinguished from other surgical instruments and does not change dramatically when it is used. However, despite the superior performance of RASNet, there are still some surgical instruments that have poor segmentation results such as Prograsp Forceps and Grasping Retractor. The mean Dice of segmenting Prograsp Forceps is just 84.59$\%$, which is much lower than that of segmenting all surgical instruments. The mean Dice of segmenting Grasping Retractor is just 88.05$\%$. After careful analysis, we found the following reasons. Prograsp Forceps is very similar to Bipolar Forceps. So it is often misclassified as Bipolar Forceps, which results in poor performance. As for Grasping Retractor, there are very few samples of Grasping Retractor in the dataset, which leads to underfitting of the network.

\section{Conclusion}
In this paper, we propose a new network which utilizes attention mechanism for segmentation. The excellent performance of RASNet is confirmed throughout the experiment. The results also show that Attention Fusion Module can greatly improve segmentation accuracy. This provides a new direction for improving the performance of the network. In the future, we will focus on designing more effective attention modules.
\section*{ACKNOWLEDGMENT}
This research is supported by the National Natural Science Foundation of China (Grants 61533016, U1713220, U1613210), the National Key Research and Development Program of China (Grant 2017YFB1302704) and the Strategic Priority Research Program of CAS (Grant XDBS01040100).

\addtolength{\textheight}{-12cm}   

\bibliographystyle{IEEEtran}
\bibliography{myreference}

\end{document}